\documentclass{article}

\usepackage[accepted]{icml2024}

\usepackage{microtype}
\usepackage{graphicx}
\usepackage{subfigure}
\usepackage{booktabs}
\usepackage[algo2e,ruled,noend]{algorithm2e}
\usepackage{hyperref}

\usepackage{amsmath}
\usepackage{amssymb}
\usepackage{mathtools}
\usepackage{amsthm}
\usepackage[capitalize,noabbrev]{cleveref}
\usepackage{colortbl}
\usepackage{pifont}
\definecolor{lightblue}{rgb}{0.93,0.95,1.0}
\theoremstyle{plain}

\theoremstyle{definition}

\theoremstyle{remark}

\usepackage{comment}
\usepackage{booktabs} % For prettier tables
\usepackage{siunitx} % For aligning numbers
\usepackage{multirow}
\usepackage[textsize=tiny]{todonotes}

\icmltitlerunning{MLIP: Efficient Multi-Perspective Language-Image Pretraining with Exhaustive Data Utilization}

\begin{document}

\twocolumn[
\icmltitle{MLIP: Efficient Multi-Perspective Language-Image Pretraining \\with Exhaustive Data Utilization}

\icmlsetsymbol{equal}{*}

\begin{icmlauthorlist}
\icmlauthor{Yu Zhang}{tj}
\icmlauthor{Qi Zhang}{equal,tj}
\icmlauthor{Zixuan Gong}{equal,tj}
\icmlauthor{Yiwei Shi}{ub}
\icmlauthor{Yepeng Liu}{uf}
\icmlauthor{Duoqian Miao}{tj}
\\
\icmlauthor{Yang Liu}{nudt}
\icmlauthor{Ke Liu}{ncrc}
\icmlauthor{Kun Yi}{bit}
\icmlauthor{Wei Fan}{oxford}
\icmlauthor{Liang Hu}{tj}
\icmlauthor{Changwei Wang}{ia}
\end{icmlauthorlist}

\icmlaffiliation{tj}{Tongji University}
\icmlaffiliation{ub}{University of Bristol}
\icmlaffiliation{uf}{University of Florida}
\icmlaffiliation{nudt}{National University of Defense Technology}
\icmlaffiliation{ncrc}{National Clinical Research Center for Mental Disorders \& National Center for Mental Disorders, Beijing Anding Hospital, Capital Medical University}
\icmlaffiliation{bit}{Beijing Institute of Technology}
\icmlaffiliation{oxford}{University of Oxford}
\icmlaffiliation{ia}{Institute of Automation, Chinese Academy of Sciences}

\icmlkeywords{Machine Learning, ICML}

\vskip 0.3in
]

\printAffiliationsAndNotice{\icmlEqualContribution}

\begin{abstract}
Contrastive Language-Image Pretraining (CLIP) has achieved remarkable success, leading to rapid advancements in multimodal studies. However, CLIP faces a notable challenge in terms of inefficient data utilization. It relies on a single contrastive supervision for each image-text pair during representation learning, disregarding a substantial amount of valuable information that could offer richer supervision. Additionally, the retention of non-informative tokens leads to increased computational demands and time costs, particularly in CLIP's ViT image encoder. To address these issues, we propose \textbf{M}ulti-Perspective \textbf{L}anguage-\textbf{I}mage \textbf{P}retraining (\textbf{MLIP}). In MLIP, we leverage the frequency transform's sensitivity to both high and low-frequency variations, which complements the spatial domain's sensitivity limited to low-frequency variations only. By incorporating frequency transforms and token-level alignment, we expand CILP's single supervision into multi-domain and multi-level supervision, enabling a more thorough exploration of informative image features. Additionally, we introduce a token merging method guided by comprehensive semantics from the frequency and spatial domains. This allows us to merge tokens to multi-granularity tokens with a controllable compression rate to accelerate CLIP. Extensive experiments validate the effectiveness of our design.
\end{abstract}

\section{Introduction}

\begin{figure}[!t]
\label{motivation}
\vspace{0.1cm}
\includegraphics[width=1\linewidth]{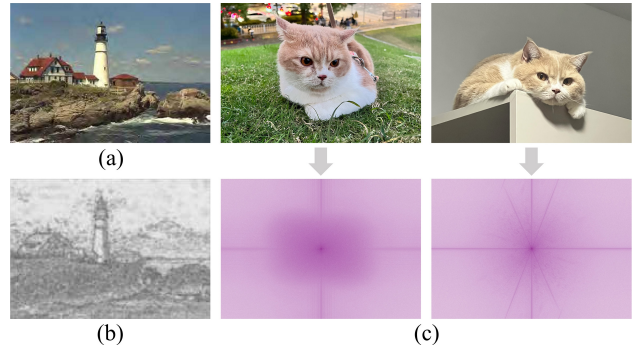}
\vspace{-0.75cm}
\caption{(a) A distorted image. (b) An objective error map. The house and the sky regions are easily observable, and those on textural regions (e.g. rocks) are less noticeable, i.e., HVS is much more sensitive to the low-frequency variations than the high-frequency variations. (c) The original images and spectrums of the same lying cat in different scenes. It shows spectrum is pretty effective in extracting and differentiating features such as the complexity and noise of a scene (the high-frequency variations).}
\vspace{-0.4cm}
\end{figure}

In recent times, multimodal study~\cite{xu2024mrftrans,zhang2024navid,xu2024local,lin2023development} has gained significant popularity, leading to rapid advancements in Vision-Language Pre-training (VLP). One noteworthy development is Contrastive Language-Image Pretraining (CLIP)~\cite{radford2021learning}, which has demonstrated remarkable performance across a range of downstream tasks. CLIP achieves this by training on 400 million image-text pairs with a contrastive mechanism to effectively bring the representations of intra-pairs closer together while pushing apart those of inter-pairs. However, upon closer examination, it is evident that CLIP encounters a significant obstacle in terms of inefficient data utilization. For example, only one contrastive supervision is utilized for each pair during the forward process, thereby leaving substantial uni-modal and cross-modal information untapped, which can potentially enhance representation. Additionally, the presence of non-informative tokens leads to increased computational requirements and time costs, especially in CLIP's image encoder, i.e., a Vision Transformer (ViT)~\cite{dosovitskiy2020image}. This additional workload hampers cross-modality alignment and significantly slows down the overall training speed of CLIP.

There has been a surge in studies focusing on the development of data-efficient CLIP-like models. Prevailing approaches include self-supervision or image enhancement techniques to increase the diversity of supervision~\cite{mu2022slip,li2021supervision,lee2022uniclip} or probe token-level alignments to refine feature learning~\cite{yao2021filip,zou2022tokenflow}.  Although these approaches have shown promising results, they primarily focus on enhancing feature learning in the single spatial domain. However, it is crucial to recognize that a 2D image signal contains a wealth of additional important features that can be extracted in the frequency domain. CNNs and ViTs, which primarily operate in the spatial domain, are devised to mimic the human visual system (HVS)~\cite{kim2017deep}. However, HVS exhibits varying sensitivity to different frequency components, as illustrated in Figure \ref{motivation}(a) and (b). Fortunately, frequency transformation techniques can naturally differentiate and isolate less sensitive frequency components, cf. Figure \ref{motivation}(c). This persuades us that image signals in the frequency domain offer valuable information, potentially promoting multi-domain supervision to enhance data efficiency.

\begin{figure}[htb]
\vspace{0.1cm}
\centering
\includegraphics[width=1.01\linewidth]{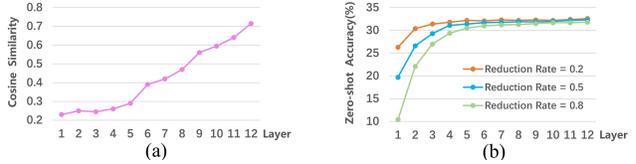}
\vspace{-0.75cm}
\caption{The observation of performing similarity calculation and reduction operations on tokens on ImageNet. (a) Average token similarity in each layer of CLIP-ViT-B/32. (b) Zero-shot accuracy of random token reductions on different layers.}
\label{reduce}
\vspace{-0.4cm}
\end{figure}

Moreover, the additional merits of frequency analysis, including computation efficiency, energy compacting, and a global view~\cite{abs-2311-06184,abs-2302-02173}, further encourage us to contemplate data utilization from multiple perspectives to improve CLIP accuracy and efficiency. i) \textit{Multi-domain}: how to benefit from complementary supervision in the frequency domain to enhance the spatial domain? ii) \textit{Multi-level}: How to introduce token-level alignments to promote instance-level alignments with fine-grained representation learning? Note that there is a fundamental distinction between frequency tokens and spatial tokens: Frequency tokens are numerous in quantity but in relatively low-frequency semantics, while spatial tokens are reduced in number, akin to text tokens, but carry high-frequency semantics. The distinction necessitates the multi-level alignments to alleviate semantics mismatch and suggests a strict one-to-one alignment at the spatial token level and a loose one-to-many alignment at the token-level alignment with text, respectively. iii) \textit{Multi-granularity}: Does global frequency effectively guide to merge tokens at multi-granularity tokens for computational efficiency? Existing studies to accelerating CLIP, like FLIP~\cite{li2023scaling} and A-CLIP~\cite{yang2023attentive}, utilize masking image patches to achieve reduced accelerating CLIP in ViT, however, suffering from information loss from unreliable masking~\cite{liang2022not}. Figure \ref{reduce}(a) and (b) show that tokens exhibit higher semantic similarity in deeper layers where reducing the token number should be more reliable. Frequency information, especially high frequency, effectively complements HVS with comprehensive semantics, potentially enhancing reliable token merging.

In light of the above discussion, we present a novel \textbf{M}ulti-Perspective \textbf{L}anguage-\textbf{I}mage \textbf{P}retraining (\textbf{MLIP}). The approach utilizes frequency analysis to ground the model and offers joint spatial-frequency token alignment, enabling exhaustive data utilization. Specifically, we propose splitting the image encoder into two stages: the Frequency Stage and the Spatial Stage, to provide frequency and spatial features of images, respectively. The Frequency Stage leverages the Discrete Fourier Transform (DFT) to efficiently mix tokens, allowing the image encoder to capture high-frequency variation features such as textures. On the other hand, the Spatial Stage utilizes the attention mechanism to learn local or global spatial features, including shape and position. These two stages generate a frequency embedding and a spatial embedding for each image, which are then used in contrastive learning alongside text embeddings at the instance level. Additionally, MLIP aligns tokens from the Frequency Stage and Spatial Stage with the text tokens at the token level, employing a loose one-to-one and a strict one-to-many matching respectively to fine-grain representation learning. To accelerate MLIP, we employ a token merging method guided by frequency-spatial supervision to reduce the token number at a controlled compression rate. Briefly, we devise a light \textit{Guide} module to process the low-resolution counterpart of the image and send its class token and global features to cross-attention layers of the Spatial Stage, to enhance image tokens. This injection of high-level semantic information provides additional reliable guidance for selecting similar image tokens to merge.

Our contributions can be summarized as follows:
\vspace{-0.2cm}
\begin{itemize}
    \vspace{-0.1cm}
    \item Multi-domain supervision: We introduce the frequency transform into the CLIP paradigm for the first time, breaking the previous practice of only mining information from image-text pairs in the spatial domain. This enables a more thorough exploration of image features, leading to a more powerful CLIP paradigm.
    \vspace{-0.1cm}
    \item Merging leads to acceleration: We utilize token merging to reduce the token number while maintaining information integrity. As far as we know, this is the first application of token merging in the CLIP paradigm.
    \vspace{-0.1cm}
    \item Multi-perspective optimization: MLIP optimizes CLIP from multi-domain, multi-level and multi-granularity perspectives. Extensive experiments validate the effectiveness of our methods, demonstrating efficiency in both data utilization and model training.
\end{itemize}
\vspace{-0.2cm}

\begin{figure*}[!t]
\includegraphics[width=1\linewidth]{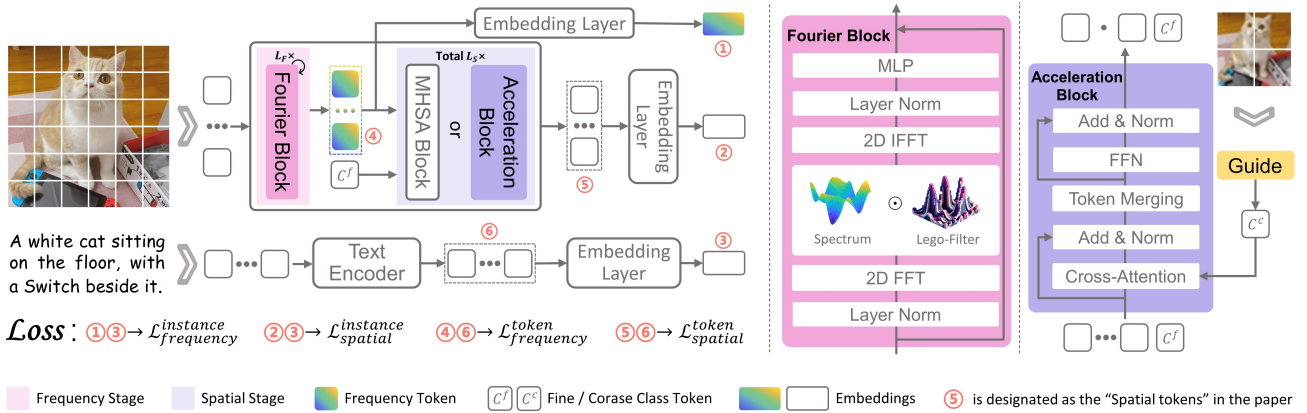}
\vspace{-0.6cm}
\caption{The overall framework of MLIP. We modify the image encoder, and related design lies in the colorful areas and indexes.}\label{framework}
\vspace{-0.3cm}
\end{figure*}

\section{Related Work}
CLIP is a simple yet powerful paradigm of representation learning. 
It is widely applied to various downstream tasks~\cite{wan2024med,liu2024zeroshot,gong2024litemind}. However, CLIP is in inefficient data utilization. Several studies are attempting to address the issue. For instance, SLIP~\cite{mu2022slip} and DeCLIP~\cite{li2021supervision} expand contrastive supervision; FILIP~\cite{yao2021filip} explores token-level alignment; CLIP-PSD~\cite{andonian2022robust} and SoftCLIP~\cite{gao2023softclip} soften one-hot labels; FLIP~\cite{li2023scaling} achieves acceleration by randomly masking patches, and A-CLIP~\cite{yang2023attentive} further masks patches with weak semantic correlation to speed up. Recent studies combine various aforementioned techniques to explore new approaches~\cite{gao2022pyramidclip,yang2023alip,dong2023maskclip,geng2023hiclip}. Diverging from the above works, we solve the issue from new perspectives: frequency transforming and token merging.

Frequency transforming plays a crucial role in signal processing and has shown surprising performance when applied to various fields of deep learning~\cite{zheng2021learning,cao2020spectral,qin2021fcanet}. These studies utilize Fourier Transform (FT) in converting signals from the spatial or time domain to the frequency domain. FNet~\cite{lee2021fnet}, as the first work to explore the application of frequency transforming to Transformer~\cite{vaswani2017attention}, finds that FT can replace the Self-Attention layer to achieve fast token mixing. GFNet~\cite{rao2021global}, applying Fast Fourier Transform (FFT) to ViT, improves the image classification performance of ViT. Subsequent studies, such as AFNO~\cite{guibas2021adaptive} and AFFNet~\cite{huang2023adaptive}, delve deeper into the application of FFT to ViT. Currently, frequency transforming is rarely discussed in VLP. We consider introducing frequency transforming in CLIP to achieve efficiency. However, unlike above works, we use both frequency and spatial domain features of images and have made modifications to the process of transforming.

\section{Methodology}
In this section, we first introduce some CLIP preliminaries and the overall MLIP loss. Sequentially present our three methods: supervision expansion via frequency transforming, joint spatial-frequency token alignment, and acceleration via token merging. Figure  \ref{framework} shows the overall framework.

\subsection{CLIP Preliminaries and MLIP Overall Loss}
For a batch of $N$ image-text pairs $\{(I_j, T_j)\}_{j=1}^{N}$, $I_j$ and $T_j$ are the image and text of the $j$-th pair. $y_j$ and $z_j$ represent the normalized embeddings of $I_j$ and $T_j$, respectively, obtained from the image encoder and text encoder. The InfoNCE loss~\cite{oord2018representation} is used for contrastive learning, and the loss for image-to-text can be computed as:
\vspace{0cm}
\begin{equation}
\mathcal{L}_{IT} =\frac{1}{N} \sum_{j=1}^{N} \log \frac{\exp(sim(y_j, z_k)/\tau)}{\sum_{k=1}^{N} \exp(sim(y_j, z_k)/\tau)},\label{infoloss}
\end{equation}
where $\tau$ is a learnable temperature hyper-parameter, it is typically set to 0.07. The function $sim(,)$ is used to compute the similarity by dot product, and the text-to-image loss $L_{TI}$ can be obtained as Equation \ref{infoloss}. Therefore, the overall loss of CLIP is calculated through $\mathcal{L}_{CLIP}=\frac{1}{2}\mathcal{L}_{IT}+\frac{1}{2}\mathcal{L}_{TI}$.

Similarly, the overall loss of MLIP is donated as:
\begin{equation}
\mathcal{L}_{MLIP}=\alpha\mathcal{L}^{ins}_{fre}+\beta\mathcal{L}^{ins}_{spa}+\gamma\mathcal{L}^{tok}_{fre}+\delta\mathcal{L}^{tok}_{spa},
\end{equation}
where $ins$ and $tok$ respectively represent instance-level alignment and token-level alignment, while $fre$ and $spa$ refer to aligning text tokens with frequency tokens and spatial tokens of the image, respectively.
We set mixing coefficients $\alpha$,~$\beta$,~$\gamma$ and  $\delta$ to balance multiple losses.

\subsection{\mbox{Supervision Expansion via Frequency Transforming}}\label{s1}
\textbf{Frequency Stage.} Frequency Stage contains $L_F\times$ Fourier Blocks for transforming tokens into the frequency domain for mixing. For an image $Y$ with the resolution of $H\times W$, we first split it into $h\times w$ non-overlapping patches, $h$ and $w$ represent the number of patches split in the $H$ and $W$ directions, respectively. After patch embedding, the collection of these $C$-dimensional tokens, which serve as the input to the image encoder, is denoted as $\mathbf{y}(p,q), 1\leq p\leq h, 1\leq q\leq w$.

The spectrum is the representation of a signal in the frequency domain. Therefore, to process a discrete signal in the frequency domain, it's essential first to obtain its spectrum through Discrete Fourier Transform (DFT). The separability of 2D DFT indicates that, for a given 2D image signal $f(m,n)$, $1\leq m\leq \mathcal{M}$, $1\leq n\leq \mathcal{N}$, its 2D DFT can be separated into two 1D DFTs: first perform a 1D DFT of length $\mathcal{N}$ along one dimension of the variable $n$, then take the computed result and perform a 1D DFT of length $\mathcal{M}$ along the other dimension of the variable $m$ to obtain the spectrum $F^{\textit{2D}}(u, v)$ of the 2D signal:
\begin{align}
F^{\textit{1D}}(m, v) &= \sum_{n=1}^{\mathcal{N}} f(m, n) e^{-i2\pi vn/\mathcal{N}}, \\
F^{\textit{2D}}(u, v) &= \sum_{m=1}^{\mathcal{M}} F^{\textit{1D}}(m, v) e^{-i2\pi um/\mathcal{M}}.
\end{align}
Further, as for $\mathbf{y}(p,q)$, $1\leq p\leq h$,$1\leq q\leq w$, we obtain:
\begin{equation}\label{2ddft}
Y(u, v) =\sum_{p=1}^{h} \sum_{q=1}^{w} \mathbf{y}(p,q) e^{-i2\pi (up/h+vq/w)},
\end{equation}
where, $i$ is the imaginary unit, and $Y(u, v)$ is the spectrum of the 2D signal at $(\omega_u, \omega_v)$.  $\omega_u = 2\pi u/h$ and $\omega_v = 2\pi v/w$ correspond to the discrete frequency components in the orthogonal dimensions. Here, we adopt the standard FFT algorithm~\cite{cooley1965algorithm} to calculate the DFT.

DFT and its inverse process are lossless. Therefore, based on the fundamental properties of DFT, given a 1D spectrum $F^{\textit{1D}}(n)$, we can reconstruct the original signal $f(n)$ by Inverse DFT (IDFT):
\begin{equation}
f(n) = \frac{1}{\mathcal{N}} \sum_{n=1}^{\mathcal{N}} F^{\textit{1D}}(v) e^{i2\pi vn/\mathcal{N}}.
\end{equation}
Consequently, We can reconstruct the original 2D signal $\mathbf{y}(p,q)$ from the 2D spectrum $Y(u, v)$:
\begin{equation}
\mathbf{y}(p,q)= \frac{1}{hw} \sum_{u=1}^{h} \sum_{v=1}^{w} Y(u, v) e^{i2\pi(up/h + vq/w)}.
\end{equation}
It is noteworthy that $\mathbf{y}(p,q)\in\mathbb{R}$. According to the fundamental properties of DFT, the spectrum $Y(u,v)$ obtained by 2D DFT is conjugate symmetric about the origin, which means $Y(u,v)= Y^*(-u,-v)$. Moreover, considering the periodicity of DFT, which states $ Y(u,v)= Y(u+P,v+q)$, one can derive that $Y(p-u,q-v)= Y^*(p,q)$. This implies that half of the spectrum $Y(u,v)$ can be used to reconstruct the complete 2D signal $\mathbf{y}(p,q)$. Therefore, we adopt a smaller equivalent spectrum $Y'(u,v)$ to replace $Y(u,v)$ for signal reconstruction:
\begin{equation}
Y'=Y(:,1:w/2).
\end{equation}
Overall, we define $\mathcal{F}(\cdot)$ as the 2D DFT, for token collection $\mathbf{y}\in\mathbb{R}^{h \times w\times C}$, the spectrum of $\mathbf{y}$ can be represented as:
\begin{equation}
Y=\mathcal{F}(\mathbf{y})\in\mathbb{C}^{h \times w\times C},
\end{equation}
where $Y$ is a complex tensor. In order to reduce computation, we take half of the spectrum $Y$, denoted as $Y’\in\mathbb{C}^{h\times\frac{w}{2}\times C}$, to effectively reconstruct the original signal. Then we introduce the Lego-Filter to modulate the spectrum to the $Y'$. We utilize $X=[x_1,x_2,...,x_\aleph]$ to represent the Lego-Filter, where $\aleph$ is the number of piece filters in the Lego-Filter:
\begin{equation}
\hat{Y} = 2\sum^{\aleph}_{j=1}\frac{1}{hw}|Y'|^2 \odot x_j cos((2j-1)\pi/2\aleph),
\end{equation}
where $\odot$ is the element-wise multiplication (Hadamard product), $|Y'|^2$ is the power spectrum of $Y'$, which smooths the spectrum, highlighting the main components of the spectrum and facilitating the subsequent learning. $cos((2j-1)\pi/2\aleph)$ compacts better energy and can aggregate the more important information in a 2D signal.

Next, we utilize IFFT $\mathcal{F}^{-1}(\cdot)$ to construct and update the token collection $\mathbf{y}$:
\begin{equation}
\mathbf{y}\leftarrow \mathcal{F}^{-1}(\hat{Y}).    
\end{equation}
Finally, the tokens within the token collection, after transforming through the Frequency Stage, are termed \textit{frequency tokens} $\mathbf{y}^{fre}$. After being embedded, frequency tokens generate an instance-level frequency embedding $y^{fre}$, which is used for alignment with the instance-level embedding $z$ coming from the text encoder.

\textbf{Spatial Stage.} Based on $L_S\times$ MHSA (Multi-Head Self-Attention)Blocks, some MHSA Blocks are replaced with Acceleration Blocks to form the Spatial Stage. Spatial Stage takes frequency tokens as input and outputs \textit{spatial tokens} $\mathbf{y}^{spa}$, also using the attention mechanism for token interaction. Similarly, spatial tokens $\mathbf{y}^{spa}$ produce an instance-level spatial embedding $y^{spa}$, which aligns with $z$.

\textbf{Instance-level alignment loss.} In MLIP, instance-level alignment losses include the alignment loss of (image) frequency-text $\mathcal{L}^{ins}_{fre}$ and (image) spatial-text $\mathcal{L}^{ins}_{spa}$. Taking $\mathcal{L}^{ins}_{fre}$ as an example, it can be represented as:
\begin{align}
&\mathcal{L}^{ins}_{fre} = \frac{1}{2N} \sum_{j=1}^{N} \log \frac{\exp(sim(y_j^{fre}, z_k)/\tau)}{\sum_{k=1}^{N} \exp(sim(y_j^{fre}, z_k)/\tau)}\nonumber\\
& \quad\quad+\frac{1}{2N} \sum_{j=1}^{N} \log \frac{\exp(sim(z_j, y_k^{fre})/\tau)}{\sum_{k=1}^{N} \exp(sim(z_j, y_k^{fre})/\tau)}.
\end{align}

\subsection{Joint Spatial-Frequency Token Alignment}\label{s2}
CLIP only summarizes global visual and textual presentations (instance-level alignment), consequently overlooking a substantial amount of fine-grained information. However, token-level alignment can utilize fine-grained information to assist the model in learning more detailed features. In MLIP, token-level alignment is also categorized into (image) frequency-text and (image) spatial-text.

\textbf{Frequency-text.} We still utilize frequency tokens for token-level alignment of image-text, adopting the original approach~\cite{yao2021filip}. Specifically, We denote $l_1$ and $l_2$ as the number of frequency tokens $\mathbf{y}^{fre}$ and the number of non-padded tokens $z$ involved in late interaction, respectively. The corresponding embeddings are $a$ and $b$. We require that through calculating cosine similarities, each token involved in token-level alignment finds its most similar cross-modal token. For instance, for the $r$-th frequency token $\mathbf{y}^{fre}_{r}$, we compute the similarity of its embedding $a^r$ with all text token embeddings $\{b^s\}^{l_2}_{s=1}$, and select the highest one to represent the matching completion of $\mathbf{y}^{fre}_{r}$:
\begin{equation}
\max_{1 \leq s \leq l_2} \frac{a^r \cdot b^s}{\| a^r \|_2~ \| b^s \|_2}.
\end{equation}
Subsequently, we use the average of matchings to represent the token-level alignment $\varpi^{IT}$ from image to text:
\begin{equation}
\varpi^{IT}=\frac{1}{l_1}\sum^{l1}_{r=1}\frac{a^r \cdot b^{s^{IT}_{r}}}{\| a^r \|_2 ~\| b^{s^{IT}_{r}}\|_2},
\end{equation}
where $s^{IT}_{r}=\max_{1 \leq s \leq l_2} \frac{a^r \cdot b^s}{\| a^r \|_2~ \| b^s \|_2}$. Therefore, for $N$ image-text pairs, we can formulate the frequency token-level alignment loss from image to text $\mathcal{L}^{tok}_{fre-IT}$. Similarly, we can derive the corresponding loss from text to image $\mathcal{L}^{tok}_{fre-TI}$. Assigning a mixing coefficient of $1/2$ to each loss, we get the full loss $\mathcal{L}^{tok}_{fre}$, denoted as:
\begin{equation}
\mathcal{L}^{tok}_{fre}=-\frac{1}{2N}\sum^{N}_{j=1}\varpi  ^{IT}_j-\frac{1}{2N}\sum^{N}_{j=1}\varpi^{TI}_j.
\end{equation}

\textbf{Spatial-text.} Spatial tokens perform a spatial-text token-level alignment. We define the number of spatial tokens as $l_3$, and their corresponding embeddings as $\{c^t\}^{l_3}_{t=1}$. Due to token merging, many spatial tokens are the products of merging previously similar tokens, which have a higher level of semantic concepts compared to frequency tokens, and their number is close to that of test tokens. Therefore, we adopt a one-to-one alignment scheme, which means that every token, during cross-modal alignment, should not only look for the most similar token but also consider the cross-modal alignment of other intra-modal tokens to avoid conflicts. Hence, we can view this as a bipartite matching problem. Further considering that the similarity between cross-modal tokens could naturally serve as a weight, we model it as a maximum weight bipartite matching problem and solve it using the Kuhn-Munkres (KM) algorithm.

For embedding collections $\{b^s\}^{l_2}_{s=1}$ and $\{c^t\}^{l_3}_{t=1}$, set $l^* = \max(l^2, l^3)$. To explain more succinctly, we use the embedding's index number within the collection to represent itself. Therefore, we construct two sets  $S = \{1, 2, \cdots,l^*\}$ and $T = \{1, 2,\cdots,l^*\}$ to represent the embedding collections of text tokens and spatial tokens, respectively. If $l^2 < l^*$ or $l^3 < l^*$, add incremental elements until reaching $l^*$ to complete the construction. Set up a weight matrix $W\in\mathbb{R}^{l^*\times l^*}$, whose element $\mathbf{w}(s,t)$ is defined as follows:
\begin{equation}
\mathbf{w}(s,t) = 
\begin{cases} 
\quad\quad 0 & \text{if} ~s\ge l^2 ~\text{or}~ t \ge l^3 \\
\frac{b^s \cdot c^t}{\| b^s \|_2~ \| c^t \|_2} & \text{otherwise}
\end{cases}
\end{equation}
We initialize the index (denoting the embedding) : $L_b(s)=max_{t\in T}\mathbf{w}(s,t), \forall s\in S$ and $L_c(t)=0, \forall t\in T$. We introduce $M$ to record the matching scheme, setting $M[t]=-1, \forall t\in T$ and adopt $match$ to record whether a match has already been made, setting $match[s]=0, \forall s\in S$. Set $\hat{S}$, $\hat{T}$, and array $Slack[~]$ are used for adjustments and updates during matching. Algorithm \ref{km} shows the core process.

\begin{algorithm}[H]
    \caption{KM for matching text and spatial embedding }
    \label{km}
    \setcounter{AlgoLine}{1}
    
    \KwIn{$S$, $T$, $W$, initialized settings ($M$, $match$, $L_b$, $L_c$)}

    \KwOut{matching scheme $M$}
    
    %注意控制语句的大小写，不然调不出你想要的循环
    \For{$s\leftarrow 1$ \KwTo $l^*$} {
    
        $\hat{S}\leftarrow \{s\}$,~$\hat{T}\leftarrow \emptyset$
        
        \For{$t\leftarrow 1$ \KwTo $l^*$}{
            $Slack[t]\leftarrow L_b(s)+L_c(t)-\mathbf{w}(s,t)$}
        
        \While{$match[s]=0$}{
            \eIf{$\exists t \in (T \setminus \hat{T}) : L_b(s) + L_c(t) = \mathbf{w}(s,t)$}{
                \If{$M[t]=-1$}{
                    $M[t]\leftarrow s$, $match[s]\leftarrow 1$, break while}
                
                $S\leftarrow S\cup\{M[t]\}$,~$T\leftarrow T\cup\{t\}$
            }{
                $t^* \leftarrow\arg\min_{t \notin \hat{T}}Slack[t]$, $\Delta\leftarrow Slack[t^*]$
                
                {\bfseries update:} ~~$\forall j \in\hat{S}, l_b(j) \leftarrow l_b(j) - \Delta$
                
                \quad \quad \quad \quad $\forall k \in\hat{T}, l_c(k) \leftarrow l_c(k) + \Delta$
                
                \quad \quad \quad \quad$\forall t \notin\hat{T}, Slack[t] \leftarrow Slack[t] - \Delta$
            }
        }
    }
    
    \Return $M$
\end{algorithm}
\vspace{-0.2cm}
After obtaining the matching scheme $M$, we can use $M$ to calculate the final spatial-text token-level alignment loss $\mathcal{L}^{tok}_{spa}$. For $N$ image-text pairs, $\mathcal{L}^{tok}_{spa}$ can be calculated by the following equation:
\begin{equation}
\mathcal{L}^{tok}_{spa}=-\frac{1}{N min(l^2,l^3)}\sum_{j=1}^{N} \sum_{t=1}^{l^*} \mathbf{w}_j(M[t], t).
\end{equation}

\vspace{0.2cm}
\subsection{Acceleration via Token Merging}\label{s3}
\textbf{Constitution of Spatial Stage.} Accelerating training is realized in Acceleration Blocks, precisely owing to the Token Merging module within them. The key to acceleration lies in selecting tokens for merging to reduce their quantity. As shown in Figure \ref{reduce}(b), performing the same token reduction operation in different blocks leads to huge performance differences. Therefore, deciding when to reduce the token number, or in other words, how to place Acceleration Blocks in the Spatial Stage, is vital. Due to the input of the Spatial Stage, frequency tokens, do not interact in the spatial domain. To enable these tokens to capture spatial features (such as shape, position, etc.), and establish local and global relationships, we set a few MHSA Blocks in the early Spatial Stage for token interaction. The subsequent setting combines MHSA Blocks and Acceleration Blocks. Moreover, a (fine) class token $C^f$ is appended to record the importance of each token for later guiding the merging. 

\textbf{Guide and cross-attention.} Due to the computational complexity of self-attention being quadratically related to the token number, MHSA Blocks should be minimized. However, this might lead to insufficient interaction between tokens before merging. A lack of global understanding may result in suboptimal merging. To resolve this paradox, we equipped Acceleration Blocks with a \textit{Guide}. Simply put, Guide is a lightweight pre-trained ViT. We split the low-resolution counterpart of the image into fewer patches as the Guide's input, so the Guide could learn more global features of the image with less computational cost, which has higher semantic concepts. Then we input its (coarse) class token $C^c$ into the cross-attention layer, where it interacts with original image tokens and (fine) class token. This process injects higher semantics into them, allowing them to acquire more global features and directional guides during their merging.

\begin{figure}[!t]
\centering
\includegraphics[width=1\linewidth]{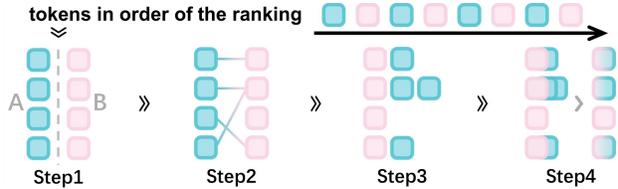}
\vspace{-0.5cm}
\caption{The process of Token Merging: Step1. Divide tokens at odd positions into set $A$ and those at even positions into set $B$. Step2. Find the most similar token in $B$ for each token in $A$ by calculating cosine similarity. Step3. Put similar tokens together to complete the match. Step4. Merge the similar tokens by weights.}\label{tome}
\vspace{-0.4cm}
\end{figure}

\textbf{Token Merging.} In the Token Merging module, we propose a controllable compression ratio token merging strategy: 1) set a compression rate $\mathcal{C}$, 2) sort all tokens from largest to smallest, based on the ranking of each token's attention score in (fine) class token $C^c$, 3) take the last $2\mathcal{C}$ tokens for merging. The merging process is shown in Figure \ref{tome}.

Our method is inspired by ToMe~\cite{bolya2022token} but differs in three main aspects: (1) Our method not only uses the fine features of the original image to determine to merge but also adds the features with higher semantics brought by Guide in the cross-attention layer to jointly determine the tokens involved in merging. (2) ToMe first performs matching, and then selects tokens for merging; while our method first selects tokens for merging, and then performs matching, this can help us achieve a controllable compression rate. (3) In our method, tokens have undergone frequency transforming, possess complementary frequency features that can better facilitate token merging.

\section{Experiment}

\subsection{Experiment Setup}
\textbf{Pre-training datasets.} To enable a fair comparison with as many methods as possible, we use YFCC15M~\cite{cui2022democratizing}, which is commonly adopted by many methods, for the pre-training of MLIP. We also adopt CC3M~\cite{sharma2018conceptual} and CC12M~\cite{changpinyo2021conceptual} for pre-training to verify that MLIP is data-efficient across multiple datasets of different scales.

\begin{table}[h]
\vspace{-0.3cm}
\centering
\caption{Comparison against CLIP baselines with zero-shot (ZS) and linear probing (LP) classification Top-1 accuracy on ImageNet,$^{\diamondsuit}$Reported in~\cite{cui2022democratizing},$^{\heartsuit}$Our implementation.}\label{result1}
\begin{center}
\begin{small}
\begin{tabular}{cccc}
\toprule
~Method~ & ~~Baesd Encoder~ &~ ZS TOP-1 ~&~ LP TOP-1 ~ ~\\
\midrule
CLIP & ViT-B / 32 & 32.8$^{\diamondsuit}$ & 62.4$^{\heartsuit }$ \\
SLIP & ViT-B / 32 & 34.3$^{\diamondsuit}$ & 67.5$^{\heartsuit }$ \\
FILIP & ViT-B / 32 & 39.5$^{\diamondsuit}$ & ~---$\quad$ \\
DeCLIP & ViT-B / 32 & 43.2$^{\diamondsuit}$ & 70.4$^{\heartsuit }$ \\
\rowcolor{lightblue}
MLIP & ViT-B / 32 & 41.1$^{~~}$  & 70.2$^{~~}$ \\
\midrule
CLIP & ViT-B / 16 & 39.0$^{\heartsuit }$  & 64.7$^{\heartsuit }$  \\
SLIP & ViT-B / 16 & 43.2$^{\heartsuit }$  & 72.3$^{\heartsuit }$  \\
DeCLIP & ViT-B / 16 & 47.9$^{\heartsuit }$  & 77.8$^{\heartsuit }$  \\
\rowcolor{lightblue}
MLIP & ViT-B / 16 & 46.3$^{~~}$  & 77.1$^{~~}$\\
\midrule
\end{tabular}
\end{small}
\end{center}
\vspace{-0.2cm}
\end{table}

\begin{table*}[t]
\centering
\vspace{-0.3cm}
\caption{Zero-shot and linear-probe classification Top-1 accuracy (\%) on 10 smaller datasets, based on variant ViT-B/32, against CLIP baselines, C10/100/F101/FLOW/SUN/DTD/CAL/AIR is CIFAR10/CIFAR-100/Food101/Flowers/SUN397/Describable Textures/Caltech-101/Aircraft. AVG is average accuracy across 10 datasets, AVG (+ImageNet) is average accuracy across 11 datasets, including ImageNet. LS denotes label smoothing. \textbf{Black text} indicates the best performance, while \underline{underlined text} indicates the second-best performance.} \label{result2}
\begin{center}
\begin{small}
\vspace{-0.3cm}
\resizebox{\linewidth}{!}{
\begin{tabular}{lcccccccccccc}
\toprule
Method & C10 & C100 & F101 & PETS & FLOW & SUN & CARS & DTD & CAL & AIR & AVG & AVG (+ImageNet) \\
\midrule
\multicolumn{13}{l}{\textit{zero-shot classification:}}\\
CLIP & 63.7 & 33.2 & 34.6 & 20.1 & 50.1 & 35.7 & 2.6 & 15.5 & 59.9 & 1.2 & 31.7 & 31.8 \\
SLIP & 50.7 & 25.5 & 33.3 & 23.5 & 49.0 & 34.7 & 2.8 & 14.4 & 59.9 & 1.7 & 29.5 & 30.0 \\
FILIP & 65.5 & 33.5 & 43.1 & 24.1 & 52.7 & 50.7 & \underline{3.3} & 24.3 & 68.8 & \textbf{3.9} & 37.0 & 37.2 \\
DeCLIP & \underline{66.7} & \underline{38.7} & \textbf{52.5} & \textbf{33.8} & 60.8 & 50.3 & \textbf{3.8} & 27.7 & 74.1 & 2.1 & 41.1 & 41.3 \\\rowcolor{lightblue}
MLIP & 65.8 & 37.0 & 48.5 & 31.7 & \underline{64.7} & \underline{52.9} & 3.0 & \underline{36.8} & \underline{75.9} & 3.1 & \underline{41.9} & \underline{41.9}\\\rowcolor{lightblue}
MLIP+LS & \textbf{67.1} & \textbf{38.9} & \underline{49.6} & \underline{32.5} & \textbf{65.3} & \textbf{53.5} & \underline{3.3} & \textbf{37.8} & \textbf{76.1} & \underline{3.2} & \textbf{42.7} & \textbf{42.8}\\
\midrule
\multicolumn{13}{l}{\textit{linear-probe classification:}}\\
CLIP & 86.5 & 64.7 & 69.2 & 64.6 & 90.6 & 66.0 & 24.9 & 61.3 & 79.1 & 23.1 & 63.0 & 63.2 \\ 
SLIP & 86.4 & 65.1  & 73.9 & 69.5  & 89.2 & 70.6 & 27.0 & 64.1 & 82.8 & 25.7 & 65.4 & 65.6 \\ 
DeCLIP & \underline{89.2} & \underline{69.0} & \textbf{75.4} & \underline{72.2} & 94.4 & 71.6 & \textbf{31.0} & 68.8 & 87.9 & \underline{27.6}& 68.7 & 68.8 \\\rowcolor{lightblue}
MLIP & 88.6 & 67.0 & 72.3 & 69.9 & \underline{96.7} & \underline{75.1} & 26.8 & \underline{83.3} & \underline{92.2} & 26.1 & \underline{69.8} &\underline{69.8} \\\rowcolor{lightblue}
MLIP+LS & \textbf{90.3} & \textbf{70.7} & \underline{73.4} & \textbf{72.5} & \textbf{97.0} & \textbf{75.6} & \underline{27.9} & \textbf{84.6} & \textbf{92.5} & \textbf{28.2} & \textbf{71.3} &\textbf{71.3} \\
\bottomrule
\end{tabular}
}
\end{small}
\end{center}
\vspace{-0.3cm}
\end{table*}

\textbf{Implementation details.} We utilize two ViT variants, ViT-B/32 and ViT-B/16, as the basis for constructing the image encoders, corresponding to MLIP-ViT-B/32 and MLIP-ViT-B/16, respectively. More details are in Appendix \ref{ioma}. The image resolution is $224\times224$, and the Guide employs DeiT-Tiny~\cite{wang2023closer} based on MAE pre-training to process the counterpart with a resolution of $64\times64$. The text encoder follows the original design of CLIP. Drawing from experience, we set the mixing coefficients $\alpha$,~$\beta$,~$\gamma$, and $\delta$ to 0.15, 0.65, 0.1, and 0.1, respectively. We train all models for 32 epochs with the same hyperparameter setting. More details are in Appendix \ref{id}.

\textbf{Downstream tasks for evaluation.} We evaluate MLIP on three downstream tasks: zero-shot and linear-probe image classification, and zero-shot image-text retrieval. For image classification, we perform experiments on ImageNet~\cite{deng2009imagenet} and 10 other smaller datasets, a total of 11 datasets. For image-text retrieval, we set experiments on Flickr30K~\cite{hodosh2013framing} and MS-COCO~\cite{chen2015microsoft}. More information is in Appendix \ref{iod}.

\subsection{Main Results}
\textbf{Zero-shot and linear-probe image classification.} Zero-shot and linear-probe classification results of MLIP on ImageNet are shown in Table \ref{result1}. It can be seen that MLIP's accuracy surpasses that of CLIP and CLIP-like baselines. However, it's worth noting that DeCLIP outperforms our MLIP, primarily because Nearest-Neighbor Supervision (essentially a kind of label smoothing) significantly enhances performance, and \cite{gao2023softclip} holds the same view. Therefore, we further conduct experiments with label smoothing and find that MLIP could exceed the performance of DeCLIP. More details are in Section \ref{ablation}. In Table \ref{result2}, we also present the zero-shot and linear-probe classification results on other datasets, where our MLIP is still competitive overall. Especially when combined with label smoothing, MLIP demonstrates a significant performance advantage. Notably, on datasets like FLOW and DTD, which contain more scenes, edges, and textures, MLIP's superiority is particularly evident. This aligns well with our expectations when introducing frequency domain transformation.

\textbf{Zero-shot image-text retrieval.}
We evaluate MLIP's zero-shot image-text retrieval performance in Table \ref{result4}, indicating that MLIP outperforms CLIP or CLIP-like methods. We can find notable improvements in most recall@1 metrics, which we attribute to the increased supervision and finer alignment. Additionally, unlike classification, image-text retrieval involves processing more complex and noisy image information, such as scene details, an area where frequency transforming excels, hence yielding better results.

\subsection{Ablation Study}\label{ablation}
In this section, we investigate the effectiveness of every design in MLIP. Unless specifically stated, all experiments use the MLIP-ViT-B/16 model pre-trained for 25 epochs on CC3M to evaluate its zero-shot classification on ImageNet and zero-shot image-text retrieval on MS-COCO.

\textbf{Data efficiency across multiple datasets of different scales.} To confirm MLIP's data efficiency across varying dataset scales, we test its performance on various pretraining datasets and datasets with fewer image-text pairs. Table \ref{result8} shows MLIP's data efficiency on different scales.

\begin{table}[h]
\vspace{-0.5cm}
\centering
\caption{Data efficiency experiments on datasets of different scales.}\label{result8}
\begin{center}
\vspace{-0cm}
\begin{small}
\begin{tabular}{cccc}
\toprule
Method &Pretraing Dataset&Baesd Encoder& ZS TOP-1\\
\midrule
~~CLIP$^{\heartsuit}$& CC3M&ViT-B / 16&16.3\\
\rowcolor{lightblue}MLIP& CC3M &ViT-B / 16&\textbf{18.4(+2.1)}\\\rowcolor{lightblue}MLIP& CC3M(95\%) &ViT-B / 16&17.9(+1.6)\\
\midrule
~~CLIP$^{\heartsuit}$& CC12M &ViT-B / 16&30.4\\
\rowcolor{lightblue}MLIP& CC12M &ViT-B / 16&\textbf{33.2(+2.8)}\\\rowcolor{lightblue}MLIP& CC12M(90\%) &ViT-B / 16&31.1(+0.7)\\
\midrule
~~CLIP$^{\heartsuit}$& YFCC15M &ViT-B / 16&37.6\\
\rowcolor{lightblue}MLIP& YFCC15M &ViT-B / 16&\textbf{42.9(+5.3)}\\\rowcolor{lightblue}MLIP& YFCC15M(90\%) &ViT-B / 16&39.4(+1.8)\\
\bottomrule
\end{tabular}
\end{small}
\end{center}

\end{table}

\begin{table*}[t]
\vspace{-0.3cm}
\centering
\begin{small}
\caption{Zero-shot image-text retrieval results on Flickr30k and MS-COCO.}\label{result4}
\vspace{0.1cm}
\resizebox{\linewidth}{!}{
\begin{tabular}{lcccccccccccc}
\toprule
&\multicolumn{6}{c}{Image-to-text retrieval} & \multicolumn{6}{c}{Text-to-image retrieval}\\
&\multicolumn{3}{c}{Flickr30k} & \multicolumn{3}{c}{MS-COCO} & \multicolumn{3}{c}{Flickr30k} & \multicolumn{3}{c}{MS-COCO} \\
Method & {R@1} & {R@5} & {R@10} & {R@1} & {R@5} & {R@10} & {R@1} & {R@5} & {R@10} & {R@1} & {R@5} & {R@10} \\
\midrule
CLIP-ViT-B / 32 & 34.9 & 63.9 & 75.9 & 20.8 & 43.9 & 55.7 & 23.4 & 47.2 & 58.9 & 13.0 & 31.7 &42.7\\
SLIP-ViT-B / 32 & 47.8 & 76.5 & 85.9 & 27.7 & 52.6 & 63.9 & 32.3 & 58.7 & 58.8 & 18.2 & 39.2 & 51.0  \\
DeCLIP-ViT-B / 32 & 51.4 & 80.2 & 88.9 & 28.3 & 53.2 & 64.5 & 34.3 & 60.3 & 70.7 & 18.4 & 39.6 & 51.4 \\
UniCLIP-ViT-B / 32& 52.3 & 81.6 & 89.0 & 32.0 & 57.7 & 69.2 & 34.8 & 62.0 & 72.0 & 20.2 & 43.2 & 54.4 \\
\rowcolor{lightblue} MLIP-ViT-B / 32 & \textbf{53.1} & \textbf{84.0}& \textbf{93.8} & \textbf{32.6} & \textbf{59.1} &\textbf{71.3} & \textbf{35.2} &  \textbf{62.9}  & \textbf{74.7} & \textbf{20.4} & \textbf{43.7} &  \textbf{56.2}\\
\midrule\rowcolor{lightblue}
MLIP-ViT-B / 16 & 56.5 & 88.7 & 98.5 & 34.7 & 64.4 & 75.8 & 37.3 & 64.8 & 78.1 & 24.5& 47.7& 62.1\\
\bottomrule
\end{tabular}
}
\end{small}
\vspace{-0.4cm}
\end{table*}

\textbf{Influence of label softening.} We employ Nearest-Neighbor Supervision (NNS)~\cite{li2021supervision} and the label smoothing method (LS) from PyramidCLIP~\cite{gao2022pyramidclip} to explore how label softening enhances MLIP's performance. As shown in Table \ref{ls}, we can see that label softening significantly boosts MLIP's performance. Therefore, without the label softening trick, MLIP outperforms DeCLIP.

\begin{table}[h]
\vspace{-0.5cm}
\centering
\caption{The influence of label softening on MLIP's performance.}\label{ls}
\begin{center}
\vspace{-0cm}
\begin{tabular}{lcc}
\toprule
Method &~~ZS TOP-1~~&~~LP TOP-1~~\\
\midrule
DeCLIP-ViT-B/32& 43.2& 70.4\\
\rowcolor{lightblue}MLIP-ViT-B/32& 41.1 &70.2\\
\rowcolor{lightblue}MLIP-ViT-B/32 + NNS& 43.1 &71.6\\
\rowcolor{lightblue}MLIP-ViT-B/32 + LS& \textbf{43.6} &\textbf{71.7}\\
\midrule
DeCLIP-ViT-B/16&47.9&77.8\\
\rowcolor{lightblue}MLIP-ViT-B/16&46.3&77.1\\
\rowcolor{lightblue}MLIP-ViT-B/16 + NNS& 48.2&78.5\\
\rowcolor{lightblue}MLIP-ViT-B/16 + LS& \textbf{48.6}&\textbf{78.9}\\
\bottomrule
\end{tabular}
\end{center}
\vspace{-0.3cm}
\end{table}

\textbf{Effectiveness of frequency transforming and token-level alignment.} To verify the effectiveness of these methods, we conduct an ablation experiment as shown in Table \ref{result5}. We can observe that both frequency transforming and token-level alignment markedly enhance the performance. This indicates that expanding the supervision of learning representation through these two methods is quite effective.

\begin{table}[h]
\vspace{-0.5cm}
\centering
\caption{Ablation study on the effects of frequency transforming (Fre-T) and token-level alignment (Tok-A). \textquotesingle{}I2T\textquotesingle{} and \textquotesingle{}T2I\textquotesingle{} mean image-to-text and text-to-image, respectively. $^1$Only use the one-to-many matching strategy. $^2$Use both one-to-many and one-to-one matching strategies, i.e., our token-level alignment method.}\label{result5}
\begin{center}
\vspace{-0.3cm}
\begin{small}
\resizebox{\linewidth}{!}{
\begin{tabular}{lccc}
\toprule
Method & I2T R@1 & T2I R@1 & ZS TOP-1 \\
\midrule
CLIP (ViT-B / 16) & 10.4 & 6.6 & 16.3 \\
~+ Fre-T& 12.7 & 8.0 & 17.6 \\
~+ Tok-A$^1$& 13.1 & 8.2 & 17.9 \\
\rowcolor{lightblue}
~+ Fre-T + Tok-A$^2$& \textbf{14.8} & \textbf{9.5} & \textbf{18.7} \\
\bottomrule
\end{tabular}
}
\end{small}
\end{center}
\vspace{-0.3cm}
\end{table}

\textbf{Influence of matching strategies.} We design a set of experiments to analyze the influence of using different matching strategies on performance. As shown in Table \ref{result6}, the MLIP’s matching strategy realizes the best result, underscoring its importance in token-level alignment. 

\begin{table}[htb]
\vspace{-0.2cm}
\centering
\caption{Ablation study on the influence of matching strategies. \textquotesingle{}o-to-o\textquotesingle{} and \textquotesingle{}o-to-m\textquotesingle{} are one-to-one and one-to-many, respectively. }\label{result6}
\vspace{0cm}
\begin{center}
\begin{small}
\begin{tabular}{ccccc}
\toprule
Fre-Text & Spa-Text & I2T R@1 & T2I R@1 & ZS TOP-1 \\
\midrule
o-to-o&o-to-m& 13.0 & 8.2 & 17.8 \\
o-to-o&o-to-o& 13.4 & 8.5 & 18.0 \\
o-to-m&o-to-m& 14.1 & 8.9 & 18.3 \\
\rowcolor{lightblue}
o-to-m&o-to-o& \textbf{14.6} & \textbf{9.3} & \textbf{18.4} \\
\bottomrule
\end{tabular}
\end{small}
\end{center}
\vspace{-0.6cm}
\end{table}

\textbf{Effectiveness of Guide.} Table \ref{result7} shows the results with and without Guide, demonstrating that Guide is essential for better performance. Furthermore, combining data from Table \ref{result5}, it's evident that the performance loss due to token merging operations can be largely compensated for by Guide. Additionally, even when compared to other baselines without the pre-trained Guide, MLIP remains competitive. This indicates that MLIP's performance gains are also attributed to other well-designed components beyond Guide.

\begin{table}[h]
\vspace{-0.5cm}
\caption{Ablation study on the effectiveness of Guide.}\label{result7}
\vspace{0.15cm}
\begin{center}
\begin{small}
\begin{tabular}{lccc}
\toprule
Method & ~I2T R@1~ & ~T2I R@1~ & ~ZS TOP-1~ \\
\midrule
CLIP& 14.1 & 8.9 & 18.0 \\
SLIP& 14.1 & 8.9 & 18.0 \\
DeCLIP w/o NNS& 14.1 & 8.9 & 18.0 \\
\rowcolor{lightblue}MLIP w/o Guide& 14.1 & 8.9 & 18.0 \\
\rowcolor{lightblue}MLIP & \textbf{14.6} & \textbf{9.3} & \textbf{18.4} \\
\bottomrule
\end{tabular}
\end{small}
\end{center}
\vspace{-0.3cm}
\end{table}
\vspace{0cm}

\textbf{Comparison of computational efficiency.} Since MLIP also reduces the computational cost by token merging, we compare its balance of performance and computation with other CLIP-like models, as shown in Table \ref{result9}. We measure the amount of computation required for each model by metric GFLOPs. In the experiment, despite MLIP obtaining sub-optimal performance, it achieves the best computation-performance balance, suggesting MLIP’s efficiency is more comprehensive.

\begin{figure*}[t]
\centering
\includegraphics[width=1\linewidth]{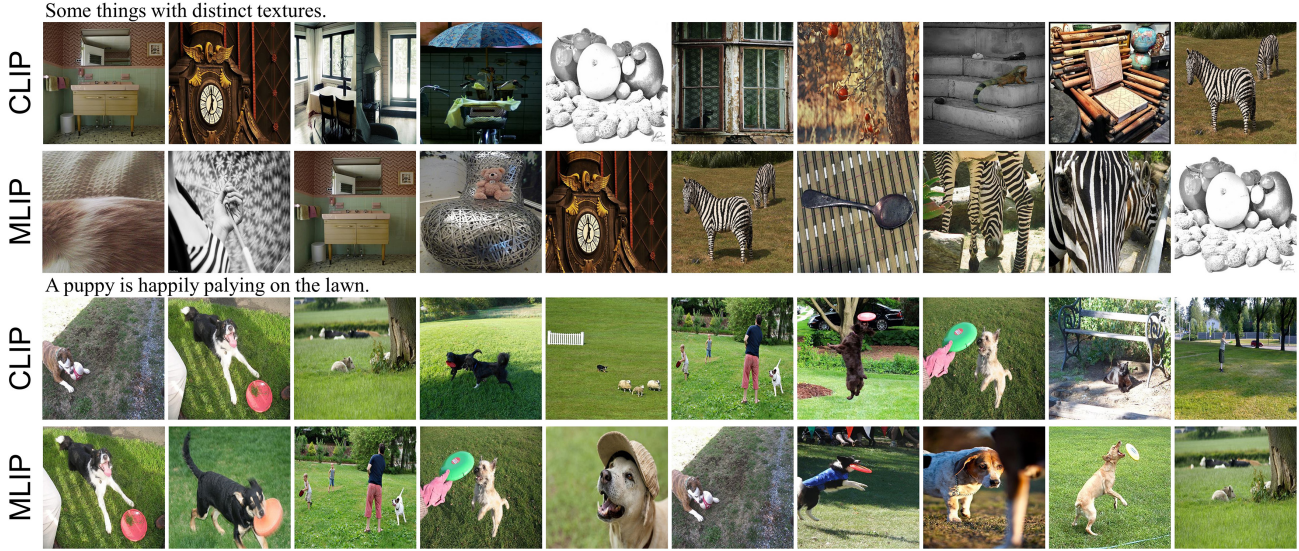}
\vspace{-0.5cm}
\caption{Text-to-image top 10 retrieval results on MS-COCO.}\label{jiansuo}
\vspace{-0.3cm}
\end{figure*}

\begin{table}[h]
\vspace{-0.5cm}
\centering
\caption{Comparison on computational efficiency. Metric ZS TOP-1/GFLOPs is used to represent computation-performance balance.}\label{result9}
\begin{center}
\vspace{-0.4cm}
\begin{small}
\resizebox{\linewidth}{!}{
\begin{tabular}{lccc}
\toprule
Method & ZS TOP-1 & GFLOPs & ZS TOP-1/GFLOPs \\
\midrule
CLIP$^{\heartsuit}$ & 16.3 & 19.78& 0.82 \\
SLIP$^{\heartsuit}$& 16.9 & 22.61 & 0.74 \\
DeCLIP$^{\heartsuit}$ & \textbf{18.7} & 26.29 &0.71 \\
\rowcolor{lightblue}MLIP& 18.4 & \textbf{19.54} & \textbf{0.94} \\
\bottomrule
\end{tabular}
}
\end{small}
\end{center}
\vspace{-0cm}
\end{table}

\textbf{Comparison of training wall-clock time.} Reporting the wall-clock time for training more effectively demonstrates computational efficiency. Table \ref{time} reports the wall-clock times for training MLIP, CLIP, DeCLIP, SLIP. Although the overall training process of MLIP is about 17 minutes slower than that of the fastest CLIP in terms of Wall-clock time, the GPU Hours are still fewer. Given the better performance of MLIP, we  compare its balance of performance and training time with other CLIP-like models, it is evident that MLIP achieves the best balance between training time and performance.

\begin{table}[h]
\vspace{-0.5cm}
\centering
\caption{Comparison on training wall-clock time.  Metric ZS TOP-1/WCT (wall-clock time) represents training-performance balance.}\label{time}
\begin{center}
\vspace{-0.3cm}
\begin{small}
\resizebox{\linewidth}{!}{
\begin{tabular}{lccc}
\toprule
Method & ZS TOP-1 & GPU Hours & ZS TOP-1/WCT \\
\midrule
CLIP & 16.3 & \textbf{361} & 1.44 \\
SLIP & 16.9 & 488 & 1.11 \\
DeCLIP & \textbf{18.7} & 545 &1.09 \\
\rowcolor{lightblue}MLIP& 370 & \textbf{1.59} & \textbf{0.94} \\
\bottomrule
\end{tabular}
}
\end{small}
\end{center}
\vspace{-0.4cm}
\end{table}

\subsection{Visualization}
\textbf{Text-to-image retrieval.} Figure \ref{jiansuo} shows 2 sets of top 10 retrieval results on MS-COCO. In the first set, it can be seen that MLIP has a more global and distinct ability to recognize textures. And in another, MLIP also shows better retrieval. 

\begin{figure}[h]
\vspace{-0.1cm}
\centering
\includegraphics[width=1.01\linewidth]{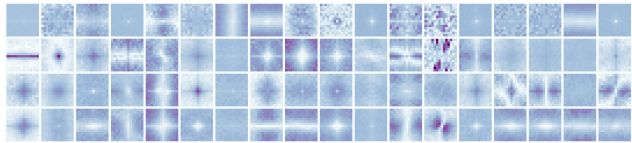}
\vspace{-0.6cm}
\caption{Visualization of Lego-Filter.}\label{filter}
\vspace{-0.2cm}
\end{figure}

\textbf{Lego-Filter.} Figure \ref{filter} visualizes Lego-Filter of MLIP-ViT-B/32, showing that Lego-Filter effectively captures both high and low-frequency variations.

\begin{figure}[h]
\vspace{-0.2cm}
\centering
\includegraphics[width=1.0\linewidth]{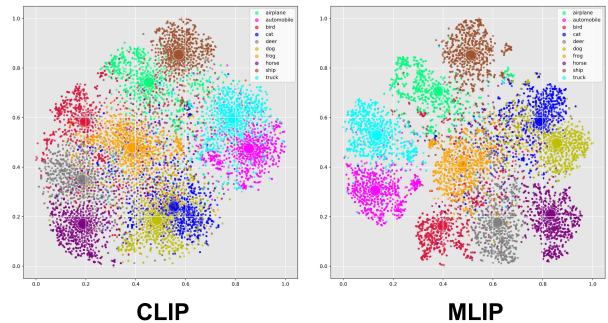}
\vspace{-0.7cm}
\caption{Visualization of embedding space.}\label{tsne}
\vspace{-0.3cm}
\end{figure}

\textbf{Embedding space.} We utilize t-SNE visualization to compare the embedding spaces of CLIP and MLIP on the CIFAR-10 dataset. From Figure \ref{tsne}, it's evident that MLIP, on the same pretrain dataset, exhibits better separation between samples of different classes, indicating that MLIP indeed improves data utilization and learns better representations.

\section{Conclusion}
In this article, we propose MLIP, a framework to develop an efficient CLIP via exhaustive data utilization in multi-perspective. MLIP introduces frequency transforming and alignments at both the token level and instance level to expand the supervision of learning representation in the image encoder. Additionally, MLIP also incorporates a modified token merging method, reducing the token number in the image encoder and accelerating the overall training. Extensive experiments validate the effectiveness of our design, and we hope our work can inspire its community. We also discuss the limitation of MLIP in Appendix \ref{limit}.

\newpage
\section*{Acknowledgements}
This work is supported by the National Key Research and Development Program of China (No. 2022YFB3104700), the National Natural Science Foundation of China (No. 62376198, No. 62006172, No. 62106091, No. 62076182 and No. 62163016), the Shandong Provincial Natural Science Foundation (No. ZR2021MF054), the Jiangxi “Double Thousand Plan” and the Jiangxi Provincial Natural Science Foundation (No. 20212ACB202001). The authors would like to thank Jun Wang, Xuerong Zhao, Yayue Tan and Jiu for inspirational suggestions and  helpful assistance.

\section*{Impact Statement}
This paper presents work whose goal is to advance the field of Machine Learning. There are many potential societal consequences of our work, none which we feel must be specifically highlighted here.

\nocite{langley00}
\bibliography{example_paper}
\bibliographystyle{icml2024}

\newpage
\appendix
\onecolumn
\section{Additional Fourier Theory Analysis}
\subsection{Discrete Fourier Transform}
The Discrete Fourier Transform (DFT) can be understood through various approaches. In this context, we explore how DFT is developed from the conventional Fourier Transform (FT), which is primarily applicable to continuous signals. The FT transforms a continuous-time signal into its frequency domain representation, acting as a broader application of the Fourier series concept. In essence, the Fourier transform for a signal \(f(t)\) is defined as follows:

\begin{equation}
F^{\textit{1D}}(i\omega) = \int_{-\infty}^{\infty} f(t)e^{-i\omega t}dt = F^{\textit{1D}}[f(t)].
\end{equation}

The Inverse Fourier Transform (IFT) bears a resemblance in structure to the Fourier Transform:

\begin{equation}
f(t) = \frac{1}{2\pi} \int_{-\infty}^{\infty} F^{\textit{1D}}(i\omega)e^{i\omega t}d\omega.
\end{equation}

The equations for the FT and the IFT provide insight into the duality characteristic of the FT, which bridges the time and frequency domains. This duality principle suggests that characteristics observed in the time domain find analogous expressions in the frequency domain. Among the many attributes of the Fourier transform, several fundamental ones include the transformation of a unit impulse function \( \delta(t) \) (commonly referred to as the Dirac delta function), which is

\begin{equation}
F^{\textit{1D}}(\delta(t)) = \int_{-\infty}^{\infty} \delta(t)e^{-i\omega t}dt = \int_{0-}^{0+} \delta(t)dt= 1,
\end{equation}

and the time-shifting property:

\begin{equation}
F^{\textit{1D}}(\delta(t - t_0)) = \int_{-\infty}^{\infty} f(t - t_0)e^{-i\omega t}dt = e^{-i\omega t_0} \int_{-\infty}^{\infty} f(t)e^{-i\omega t}dt = e^{-i\omega t_0} F^{\textit{1D}}(i\omega).
\end{equation}

In practical scenarios, continuous signals are seldom directly dealt with. Instead, a common approach involves sampling the continuous signal to generate a sequence of discrete signals. This sampling process is typically carried out through a series of unit impulse functions.

\begin{equation}
f_s(t) = f(t) \sum_{n=-\infty}^{\infty} \delta(t - nT_s)= \sum_{n=-\infty}^{\infty} f(nT_s) \delta (t-nT_s),
\end{equation}

When we take the Fourier Transform (FT) of the sampled signal \(f_s(t)\) with a sampling interval \(T\) and then apply Equation (A.3) and Equation (A.4), we obtain

\begin{equation}
F_s^{\textit{1D}}(i\omega) = \sum_{n=-\infty}^{\infty} f(nT_s) e^{-in\omega T_s}.
\end{equation}

In the provided equation, it is evident that \(F_s^{\textit{1D}}(i\omega)\) exhibits periodic behavior with a fundamental period of \(2\pi/T_s\). In fact, there is always a direct correspondence between discrete signals in one domain and periodic signals in the other domain. Typically, we prefer to work with a normalized frequency, denoted as \(\omega \leftarrow \omega T_s\), which results in \(F_s^{\textit{1D}}(i\omega)\) having an exact period of \(2\pi\). We can also represent \(f(n)\) as \(f(nT_s)\), defining it as the sequence of discrete signals, and subsequently derive the Discrete-Time Fourier Transform (DTFT):

\begin{equation}
F^{\textit{1D}}(e^{i\omega}) = \sum_{n=-\infty}^{\infty} f(n) e^{-in\omega}.
\end{equation}

When the discrete signal \(f(n)\) has a finite length \(\mathcal{N}\), which is a common scenario in digital signal processing, the DTFT can be expressed as follows:

\begin{equation}
F^{\textit{1D}}(e^{i\omega}) = \sum_{n=1}^{\mathcal{N}} f(n) e^{-in\omega}.
\end{equation}
 where the non-zero terms of the discrete signal \(f(n)\) are assumed to lie in the range \([1, N]\) without loss of generality, the DTFT \(F^{\textit{1D}}(e^{i\omega})\) is indeed a continuous function of \(\omega\). You can obtain a sequence \(F^{\textit{1D}}[v]\) by sampling \(F^{\textit{1D}}(e^{i\omega})\) at discrete frequencies \(\omega_v = \frac{2\pi v}{\mathcal{N}}\), resulting in:

\begin{equation}
F^{\textit{1D}}(v) = F^{\textit{1D}}(e^{i\omega})|_{\omega=2\pi n/N}
= \sum_{n=1}^{\mathcal{N}} f[n]e^{-i(2\pi/\mathcal{N})kn},
\end{equation}

Indeed, the extension from the 1D Discrete Fourier Transform (DFT) to the 2D DFT is straightforward. The 2D DFT can be viewed as applying the 1D DFT independently to the two dimensions of the data. Specifically, the 2D DFT of a signal or image \(f(m, n)\) is given by:

\begin{equation}
F^{\textit{2D}}(u, v) = \sum_{m=1}^{\mathcal{M}} \sum_{n=1}^{\mathcal{N}} f(m, n)e^{-i2\pi \left(\frac{um}{\mathcal{M}} + \frac{vn}{\mathcal{N}}\right)},
\end{equation}

Therefore, we can obtain Equation \ref{2ddft}.

\subsection{The equivalence
between self-attention and frequency-domain computation}
\subsubsection{(Self-Attention)}

Consider an input tensor $X$ (for clearer expression, we choose $X$ different from $F$), where $x_n \in \mathbb{R}^d$ signifies the $n$-th element in the sequence, and $N$ represents the sequence's length.

\textbf{Definition 1 Self-Attention}
The mechanism of self-attention, denoted as $\text{Self-Att}: \mathbb{R}^{N \times d} \rightarrow \mathbb{R}^{N \times d}$, is articulated through a kernel integration approach as found in \cite{kovachki2021neural,guibas2021adaptive,tsai2019transformer}:

\begin{equation}
\text{Self-Att} = \text{softmax}\left(\frac{(XW_q)(XW_k)^{\top}}{\sqrt{d}}\right)XW_v
\end{equation}

Here, $K$ is defined as the softmax-normalized score array of size $N \times N$: $K = \text{softmax}\left(\frac{(XW_q)(XW_k)^{\top}}{\sqrt{d}}\right)$. The operation of self-attention is then equivalent to an asymmetric kernel $\kappa: [N] \times [N] \rightarrow \mathbb{R}^{d \times d}$, with each entry $\kappa[s, t]$ constructed as $K[s, t] \otimes W_v^{\top}$. Thus, we interpret self-attention as a sum over this kernel.

\begin{equation}
\text{Self-Att}(X)[s] = \sum_{t=1}^{N} \kappa[s, t] \cdot X[t], \quad  \forall s \in [N] 
\end{equation}

Expanding upon the notion of kernel summation, we incorporate the concept of continuous kernel integration. Within this framework, the tensor $X$ encapsulates a spatial function over the space $X = (D, \mathbb{R}^d)$, with $D$ being a subset of $\mathbb{R}^2$.

\begin{equation}
\text{Self-Att}(X)[s] = \mathcal{K}(X)(s) = \int_{D} \kappa(s, t)  X(t) \, dt, \quad \forall s \in D 
\end{equation}

For any continuous element $X$ within $D$, we define the kernel integral operator $\mathcal{K} : (\mathbb{R}^d, D) \rightarrow (D, \mathbb{R}^d)$ in the following manner:

\textbf{Definition 2 Kernel Integral}. The kernel integral operator, denoted as $\mathcal{K}$, maps pairs of the domain $D$ and the Euclidean space $\mathbb{R}^d$ into themselves, symbolically represented as $\mathcal{K} : (D, \mathbb{R}^d) \rightarrow (D, \mathbb{R}^d)$. This operator is formally defined for all $s$ within the domain $D$ as follows:

\begin{equation}
\mathcal{K}(X)(s) = \int_{D} \kappa(s,t) \cdot X(t) \, dt,  ~~\forall s \in \mathcal{D}
\end{equation}

where $\kappa$ is a continuous function that takes two arguments from the domain $D$ and returns a $d \times d$ real matrix. In the context where Green's kernel is applied, $\kappa(s,t)$ simplifies to $\kappa(s - t)$, which characterizes a specific instance of this kernel function.

\textbf{Definition 3 Frequency-Domain Analysis}

The convolution theorem, as stated by \cite{soliman1990continuous}, posits that spatial domain convolution is functionally analogous to frequency domain multiplication. Hence, for any continuous input \( X \) belonging to domain \( \mathcal{D} \), the kernel integration as outlined by \cite{guibas2021adaptive} can be expressed as:

\begin{equation}
\mathcal{K}(X)(s) = \mathcal{F}^{-1}(\mathcal{J}(k) \cdot \mathcal{F}(X))(s),~~ \forall s \in \mathcal{D}
\end{equation}

Here, the symbol \( \cdot \) denotes element-wise multiplication.

To summarize, leveraging computations in the frequency domain to restructure self-attention mechanisms offers a method that is both effective and theoretically sound. This approach also provides a theoretical foundation for the practicality and validity of our proposed method.

\section{Additional Information of Dataset}\label{iod}
\subsection{Pre-training Dataset}
\textbf{YFCC15M.} RFCC15M~\cite{radford2021learning} is a curated subset of YFCC-100M dataset \cite{thomee2016yfcc100m}. It is specifically filtered to include only those images that have English titles or descriptions. The dataset comprises 14,829,396 images, each accompanied by natural language captions.

\textbf{CC3M.} The CC3M dataset \cite{sharma2018conceptual} is a large-scale collection of over 3 million images accompanied by natural-language captions, providing a diverse resource for automatic image captioning tasks. The images and captions in CC3M are sourced from the web, particularly from the Alt-text HTML attributes of images, offering a wide array of styles and contexts.

\textbf{CC12M.} The CC12M dataset \cite{changpinyo2021cc12m} is a collection of approximately 12 million image-text pairs, significantly larger and more diverse than its predecessor, CC3M. Designed specifically for vision-and-language pre-training.

\subsection{Downstream Task Dataset}
\subsubsection{Image Classification}
In this section, we detail the ten datasets utilized for the classification task in our experiments, each of which is concisely summarized in Table \ref{table:datasets_table}.

\textbf{ImageNet.} The ImageNet dataset \cite{deng2009imagenet} consists of millions of labeled images across a wide variety of categories. It is structured according to the WordNet hierarchy, with each node of the hierarchy represented by hundreds of images. 

\textbf{Caltech-101.} The Caltech 101 dataset \cite{fei2004learning} consists of approximately 9,000 images divided into 101 distinct object classes, along with an extra category for background/clutter. The dataset encompasses a diverse collection of objects in each class, with image counts per category ranging roughly from 40 to 800. For oriented items like airplanes and motorcycles, images have been flipped to align from left to right, and images of vertically structured objects like buildings have been rotated to be off-axis aligned.

\textbf{CIFAR-10.} The CIFAR-10 \cite{krizhevsky2009learning} dataset is a collection of 60,000 color images, each 32x32 pixels in size. These images are categorized into 10 different classes, with each class representing distinct objects.

\textbf{CIFAR-100.} The CIFAR-100 \cite{krizhevsky2009learning} dataset is similar to the CIFAR-10 dataset but with a higher granularity in classification. It contains 60,000 color images, each 32x32 pixels, divided into 100 classes. Each class has 600 images, providing a more challenging and diverse dataset for image recognition tasks.

\textbf{Describable Texture.} The Describable Textures dataset \cite{cimpoi2014describing} is a specialized collection of images focused on textures. It consists of texture images that are categorized based on describable attributes, rather than the object or material they represent. The dataset includes a wide range of textures, such as patterns found in nature, fabrics, or man-made materials. Each texture is annotated with a set of human-describable attributes, like "bubbly," "cracked," or "woven.". Its emphasis on describable attributes rather than just material types allows algorithms to better understand and interpret the various characteristics that make up a texture.

\textbf{Food-101.} The Food-101 dataset \cite{bossard2014food} is specifically designed for food recognition tasks. It contains 101,000 images, divided into 101 food categories, with each category containing 1,000 images. 

\textbf{Oxford -IIIT Pets.} The Oxford -IIIT Pets \cite{parkhi2012cats} dataset contains images of pets, specifically focused on cats and dogs. It includes 37 different breeds of cats and dogs, with roughly 200 images for each breed, totaling around 7,400 images. The images are varied in terms of scale, pose, and lighting.

\textbf{Oxford Flowers.} The Oxford Flowers dataset \cite{nilsback2008automated} consists of 8,189 images, each depicting one of 102 flower species commonly found in the United Kingdom. Each class (flower species) in the dataset is represented by between 40 and 258 images, ensuring a variety of examples for each type of flower. The images in the dataset vary in terms of scale, pose, and lighting conditions, which makes the dataset challenging for algorithms to process. In addition, there are categories that have large variations within the category and several very similar categories. The dataset is visualized using isomap with shape and color features.

\textbf{SUN397.} The SUN397 dataset \cite{xiao2010sun} is a comprehensive collection of images specifically designed for scene recognition and classification tasks in computer vision. This dataset is part of the Scene UNderstanding (SUN) database, which is focused on providing a rich variety of scene categories. The SUN397 dataset contains approximately 108,000 images that span 397 different scene categories, offering a remarkably broad spectrum of environments.

\textbf{FGVC Aircraft.} The FGVC Aircraft dataset \cite{maji2013fine} is a specialized image dataset used in fine-grained visual categorization (FGVC) tasks, particularly focusing on aircraft recognition and classification. The dataset contains 10,200 images of aircraft, with 100 images for each of 102 different aircraft model variants, most of which are airplanes. 

\begin{table}[ht]
\centering
\caption{Overview of used datasets in our classification experiments.}
\begin{tabular}{lccccc}
\toprule
Dataset            & Abbreviation & Classes & Train Size & Test Size & Evaluation Metric \\
\midrule
ImageNet           & IN           & 1000    & 1,281,167  & 50,000    & accuracy          \\
Caltech-101        & CAL          & 102     & 3,060      & 6,085     & mean per class    \\
CIFAR-10           & C10          & 10      & 50,000     & 10,000    & accuracy          \\
CIFAR-100          & C100         & 100     & 50,000     & 10,000    & accuracy          \\
Describable Textures & DTD       & 47      & 3,760      & 1,880     & accuracy          \\
Food-101           & F101         & 101     & 75,750     & 25,250    & accuracy          \\
Oxford-IIIT Pets   & PETS         & 37      & 3,680      & 3,669     & mean per class    \\
Oxford Flowers 102 & FLOW         & 102     & 2,040      & 6,149     & mean per class    \\
SUN397             & SUN          & 397     & 19,850     & 19,850    & accuracy          \\
FGVC Aircraft      & AIR          & 100     & 6,667      & 3,333     & mean per class    \\
\bottomrule
\end{tabular}
\label{table:datasets_table}
\end{table}

\subsubsection{Image-Text Retrieval}
\textbf{Flickr30K.} The Flickr30K dataset \cite{hodosh2013framing} is mainly used for image captioning and visual question-answering tasks. This dataset comprises approximately 30,000 images, and each image in the Flickr30k dataset is accompanied by five different textual descriptions (captions).

\textbf{MS-COCO.} The MS-COCO dataset \cite{chen2015microsoft} contains over 200,000 images with a diverse set of everyday scenes that include complex backgrounds and a variety of objects. It is richly annotated with details and multiple object labels. 

\section{Additional Information of MLIP Architecher}\label{ioma}
We follow the same architecture design as CLIP. Table \ref{based} and Table \ref{spec} showcase the based and specific structures of the MLIP series.

\begin{table}[H]
\centering
\caption{The architecture parameters for based models of MLIP.}\label{based}
\begin{tabular}{lcccccccc}
\toprule
\multirow{2}{*}{Model} & \multicolumn{1}{c}{Embedding} & \multicolumn{1}{c}{Input}      & \multicolumn{3}{c}{Image Encoder}                                                      & \multicolumn{3}{c}{Text Encoder}                                                       \\
                       & \multicolumn{1}{c}{dimension} & \multicolumn{1}{c}{resolution} & \multicolumn{1}{c}{\#layers} & \multicolumn{1}{c}{width} & \multicolumn{1}{c}{\#heads} & \multicolumn{1}{c}{\#layers} & \multicolumn{1}{c}{width} & \multicolumn{1}{c}{\#heads} \\\bottomrule
MLIP-ViT-B&512&224$\times$ 224&12&768&12&12&512&8                             \\
MLIP-ViT-L&512&224$\times$ 224&24&1024&16&12&768&12\\
\bottomrule
\end{tabular}
\end{table}

\begin{table}[H]
\centering
\caption{The specific structural parameters of the image encoder in MLIP.}\label{spec}
\begin{tabular}{lcccc}
\toprule
Method& Frequency Stage  &  Spatial Stage & Acceleration Block&  Compression rate\\
\bottomrule
MLIP-ViT-B / 32& $[1,2,3,4]$ & $[5,\cdots ,12]$ &  $[9,11]$ &$[0.7,0.7]$  \\
MLIP-ViT-B / 16& $[1,2,3,4]$ & $[5,\cdots ,12]$ & $[9,11]$ & $[0.5,0.5]$ \\
MLIP-ViT-L / 14& $[1,\cdots ,8]$ & $[9,\cdots ,24]$& $[13,16,19,22]$ & $[0.7,0.7,0.65,0.65]$ \\
\bottomrule
\end{tabular}
\end{table}

\section{Additional Implementation Details}\label{id}
We used the AdamW optimizer~\cite{loshchilov2017decoupled}, with a weight decay rate of 0.1 for pre-training. For the first 2000 warm-up iterations, the learning rate increases linearly to the peak value, then decays to 0 following a cosine strategy~\cite{loshchilov2016sgdr}. We set the batch size to 4096 and conducted all experiments on 32 V100 GPUs. Additionally, to save GPU memory, we use automatic mixed-precision~\cite{micikevicius2017mixed} for training. Unless specifically stated, ablation studies were conducted with training 25 epochs on CC3M. Moreover, we also briefly tested the performance of MLIP-L/32 constructed based on the ViT variant of ViT-L/32, but differently, we only trained it on YFCC15M for 8 epochs.

\section{Limitation}\label{limit}

\textbf{Structure limitation.} Given that Transformers, as opposed to CNNs, can establish long-range dependencies and dominate multimodal applications, this paper only investigates MLIP based on the ViT structure.

\textbf{Experiment limitation.} Due to limited computational resources, we are unable to extend MLIP to larger-scale models such as ViT-Large for complete experimentation. However, similar works~\cite{li2021supervision,lee2022uniclip,gao2022pyramidclip,gao2023softclip,geng2023hiclip,dong2023maskclip,yang2023alip}, have not expanded to ViT-Large either.

\textbf{Comparison limitation.} Since most similar works are not open-source, and downstream tasks, pre-training datasets, and training strategies vary widely, it is challenging for us to conduct a broader fair comparison.

\textbf{Performance limitation.} To be honest, MLIP approaches but not surpass state-of-the-art (SOTA) performance because we prioritize a more comprehensive efficiency. Some experiments show that if focusing solely on enhancing performance, with more refined optimization, it might be possible to achieve SOTA, which we will pursue in our future research.

\textbf{Transfer limitation.} Due to token merging, applying the image encoder of MLIP to other dense vision downstream tasks such as segmentation poses some challenges. However, studies~\cite{li2022exploring,zhang2022segvit,zhang2023segvitv2,chen2022vision,wang2022advancing} demonstrate that for such plain ViT models, modifications can still be made to perform dense downstream tasks like detection and segmentation.

\end{document}